# Identifying Metaphor Hierarchies in a Corpus Analysis of Finance Articles

**Aaron Gerow (gerowa@tcd.ie)**
School of Computer Science & Statistics, Trinity College Dublin
College Green, Dublin 2, Ireland

**Mark T. Keane (mark.keane@ucd.ie)**
School of Computer Science & Informatics, University College Dublin
Belfield, Dublin 4, Ireland

**Abstract**

Using a corpus of over 17,000 financial news reports (involving over 10M words), we perform an analysis of the argument-distributions of the UP- and DOWN-verbs used to describe movements of indices, stocks, and shares. Using measures of the overlap in the argument distributions of these verbs and k-means clustering of their distributions, we advance evidence for the proposal that the metaphors referred to by these verbs are organised into hierarchical structures of superordinate and subordinate groups.

**Keywords:** Metaphor; corpus analysis; word meaning; semantics; concept hierarchies; grounding.

## Introduction

In recent years, significant progress has been made in deriving meaning from statistical analyses of distributions of words (e.g., Gerow & Keane, 2011a; Landauer & Dumais, 1997; Turney & Pantel, 2010; Michel et al., 2010). This distributional approach to meaning takes the view that words that occur in similar contexts tend to have similar meanings (cf. Wittgenstein, 1953) and that by analysing word usage we can get at their meanings. For example, the word co-occurrence statistics derived in Latent Semantic Analysis (LSA) seem to tell us about the structure of the lexicon, as they are good predictors of reaction times in lexical decision tasks (Landauer & Dumais, 1997). Similarly, word patterns drawn from WordNet, have been used to determine semantic relationships accurately enough to answer the multiple-choice analogy questions on the SAT college entrance test (Turney, 2006; Veale, 2004). More generally, it has been suggested that significant insights into human culture and behaviour can be derived from analysing very large corpora, like the Google Books repository (Michel et al., 2010). In this paper, we apply similar distributional analyses to understand the structure of metaphoric knowledge behind "UP" and "DOWN" verbs from a corpus of financial news reports.

Lakoff (1992; Lakoff & Johnson, 1980) has argued that our understanding of many concepts – such as emotions and mental states – are grounded in a few ubiquitous metaphors. For example, spatial metaphors that structure emotional states, such as HAPPINESS IS UP and SADNESS IS DOWN, are found in nearly all languages. Similar spatial metaphors, of the sort we examine here, seem to ground many stock-market reports. Accounts of index, stock-market, and share movements tend to converge around metaphors of *rising* and *falling*, *attack* and *retreat*, *gain* and *loss*. These concepts appear to be grounded by core metaphors that could be glossed as GOOD IS UP and BAD IS DOWN. Lakoff and Johnson (1980) have pointed to this UP-DOWN metaphor opposition as underlying accounts of wealth (WEALTH IS UP as in *high class*), the rise and fall of numbers (MORE IS UP / LESS IS DOWN) and changes in quantity (CHANGE IN QUANTITY IS WAR as in *retreating profits* and *defensive trades*).

If such UP-DOWN metaphors ground the use of the words used in finance reports, a corpus analysis should be able reveal the structure of prominent concepts in the domain. Specifically, we analyse arguments taken by UP and DOWN verbs to determine whether there is evidence of hierarchical structuring in their grounding metaphors (see Gerow & Keane, 2011a, for a separate analysis of changes in this language over time).

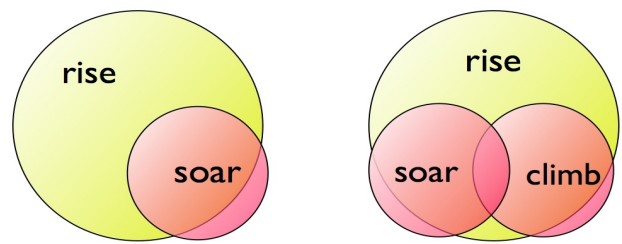

**Figure 1**: Venn Diagrams showing (**a**) *rise* covering most of *soar* and (**b**) *rise* covering *soar* and *climb*.

## Metaphor Hierarchies

We propose that if one metaphor is more fundamental than another, it should play an organizing role in structuring other concepts; specifically, it might act as a superordinate, metaphoric concept that structures some set of basic-level, metaphoric concepts (Rosch et al., 1976). So, for example, the UP metaphor schema could be a superordinate to a set of more specific metaphors (like GAIN, GROW and so on). If this is the case, then the verb that best captures this fundamental metaphor should have an argument-distribution that covers the argument-distributions of its subordinate metaphors because, by definition, it is more general. Similarly, the verbs that refer to the basic-level metaphors – what we might call *sibling metaphors* – beneath this superordinate, should have smaller, more-equal overlaps in their argument-distributions. A corollary of this proposal is that if we compare the similarity of the argument-distributions in all of these verbs, we should find the sibling metaphors clustering together separately from the superordinate metaphor-verb (see Study 2).

Figure 1 graphically illustrates these ideas using Venn diagrams. If we count the arguments taken by a given verb in a corpus, say *rise*, then each argument will have a certain frequency (see Table 1, for the 10 most frequent arguments of *rise*) giving us the argument-distribution for *rise*[1]. If we take another verb that we believe to be subordinate to *rise* – say *soar* – then it should have a more restricted scope in its arguments; furthermore, its argument distribution should be subsumed within those of *rise* (see Figure 1a). However, this coverage will be *asymmetric*; *rise*'s argument-distribution will cover that of *soar* but *soar*'s argument-distribution will not cover that of *rise*. If a third verb, say *climb*, is also subordinate to *rise* then it too might have a similar coverage pattern (see Figure 1b). Relative to one another, the two subordinates (i.e., *soar* and *climb*) could have related argument-distributions that partially overlap one another, but their respective complements will still be quite large (i.e., neither will subsume the other's argument-distribution as in Figure 1b).

Another corollary to this coverage analysis is that these patterns imply specific similarities between the argument-distributions of these verbs. *Soar* and *climb* could be quite similar (e.g., they might have 25% of the same arguments) but they will both be less similar to *rise* because *rise* has a large complement of other arguments that neither of them share. So, if we do a clustering analysis of their argument-distributions we could find that *soar* and *climb* cluster together as a group but *rise* remains separate in a (singleton) group of its own.

In summary, we take these distributional patterns of a verb's arguments as evidence of a deeper hierarchical structuring of the metaphors referred to by these words. We assume that verbs capturing superordinate metaphors will be indicated by verb distributions that:

- asymmetrically cover the argument-distributions of other (subordinate) verbs to a significant extent
- yield a distinctive singleton-grouping in clustering

In contrast, verbs that capture subordinate sibling-metaphors will:

- have a more symmetric and balanced coverage of each others' argument distributions
- be grouped together as highly similar in a clustering analysis

In this article, we report two studies examining these predictions. Study 1 looks for metaphor hierarchies in the argument-distributions of the UP verbs looking at coverage and clusters. Study 2 does the same for DOWN verbs (see Table 2). The verbs we examine were taken from the most popular verbs used to describe the nouns 'market', 'economy', 'stock', and 'price' and account for 11.1% of all verbs in the corpus. When excluding closed-class words they account for 17.8% of all verbs. We take a verb to be a verb lemma, that is, a canonical part-of-speech token derived from a single word by a probabilistic POS tagger (explained below). From these analyses, we argue for specific metaphor hierarchies in both verb sets. We conclude with a consideration of the implications of these findings. But first, consider the corpus we used for these analyses.

## The Corpus

In January, 2010, we carried out automated web searches that selected all articles referring to the three major world stock indices (Dow Jones, FTSE 100, and NIKKEI 225) from three websites: the *New York Times* (NYT, www.nyt.com), the *Financial Times* (FT, www.ft.com) and the *British Broadcasting Corporation* (BBC, www.bbc.co.uk). These searches harvested 17,713 articles containing 10,418,266 words covering a 4-year period: January 1st, 2006 to January 1st, 2010. The by-source breakdown was FT (13,286), NYT (2,425), and BBC (2,002). The by-year breakdown was 2006 (3,869), 2007 (4,704), 2008 (5,044), 2009 (3,960), and 2010 (136). The corpus included editorials, market reports, popular pieces, and technical exposés. These three resources were chosen because they are in English and have a wide-circulation and online availability. The FT made up the majority of the articles; however, the spread was actually much wider as many articles were syndicated from the Associated Press, Reuters, Bloomberg News, and Agence France-Presse. The uniqueness of the articles in the database was ensured by keying them on their first 50 characters.

**Table 1**: The percentage of *rise*'s argument distribution covered for its ten most frequent arguments ($N = 23,647$).

| Rank | Argument Word | % of Corpus |
|---|---|---|
| 1 | Index | 7.3 |
| 2 | Share | 5.6 |
| 3 | Point | 4.8 |
| 4 | Percent | 2.9 |
| 5 | Price | 2.4 |
| 6 | Stock | 2.0 |
| 7 | Yield | 1.9 |
| 8 | Cent | 1.3 |
| 9 | Profit | 0.9 |
| 10 | Rate | 0.9 |

Once retrieved, the articles were stripped of HTML, converted to UTF-8, and shallow-parsed to extract phrasal structure using a modified version of the Apple Pie Parser (Sekine, 1997). Each article was stored in a relational database with sentential parses of embedded noun- and verb-phrases. Sketch Engine was used to lemmatise and tag the corpus (Kilgarriff et al., 2004). Sketch Engine is a web-based, corpus-analysis tool that lemmatises and tags customised corpora with parts-of-speech tags using the

---
[1] Typically, we find that these distributions follow power-laws in having a few arguments that account for most of the occurrences; they will often follow a Pareto distribution, that is, 20% of the arguments account for 80% of the distribution.

TreeTagger schema (Schmid, 1994). Lemmatisation, similar to stemming, is used to reduce open-class words to a canonical, part-of-speech token (verb, noun, etc...) that includes all tenses, declensions, and pluralizations. For example, the one verb lemma "fall" includes instances such as "fall", "fell" and "falls", whereas the noun lemma "fall" includes "a fall" and "three falls". Sketch Engine provides so-called "sketches" of individual lemmas. For example, the sketch for fall-n (the word "fall" as a noun) is different from the sketch for fall-v ("fall" as a verb.) With some lemmas, the differences marked by part-of-speech are large, such as with store-n compared to store-v. These sketches facilitated the statistical analysis of the most common arguments of verbs. For example, one of the most common verbs in the corpus was "fall," which took a range of arguments with different frequencies (e.g., "DJI", "stocks", "unemployment"). Throughout this paper, when we refer verbs we take this to mean verb lemmas.

Table 2: The UP and DOWN verb-sets used in studies.

| UP-verbs | occurrences (% corpus*) | DOWN-verbs | occurrences (% corpus*) |
|---|---|---|---|
| rise | 29,261 (4.20%) | fall | 39,230 (4.20%) |
| gain | 13,134 (1.40%) | lose | 12,298 (1.30%) |
| increase | 6,158 (0.67%) | drop | 8,377 (0.90%) |
| climb | 5,631 (0.60%) | decline | 3,672 (0.39%) |
| jump | 4,960 (0.53%) | slip | 3,336 (0.36%) |
| rally | 4,190 (0.45%) | ease | 3,243 (0.35%) |
| advance | 2,385 (0.26%) | slide | 2,777 (0.30%) |
| surge | 2,313 (0.25%) | tumble | 2,135 (0.23%) |
| recover | 2,165 (0.23%) | plunge | 1,592 (0.17%) |
| soar | 1,649 (0.18%) | retreat | 1,474 (0.20%) |
| rebound | 1,220 (0.13%) | sink | 1,339 (0.14%) |
| alleviate | 134 (0.01%) | dip | 1,322 (0.14%) |
| elevate | 52 (0.00%) | worsen | 500 (0.05%) |
| | | plummet | 443 (0.05%) |
| | | decrease | 123 (0.01%) |

*Corpus $N$ = 929,735, excluding closed-class words.

We manually selected two sets of verbs from the corpus. The UP verbs were those dealing with positive, upward movements; cases like "Google *rose* rapidly today", "employment *rose*" or "stocks *climbed* to new heights". The DOWN verbs were those dealing with negative, downward movements in stocks and economic trends. Table 2 shows the two sets of verbs (and their respective coverage of all verbs in the corpus; they are arranged with their apparent antonyms).

## Study 1: UP Verbs & Metaphors

Using the corpus described in the previous section, we performed an analysis of the argument distributions of all the UP verbs. We computed two measures; their overlap and their similarity clustering. For overlap, we found the intersection of the argument-distributions of each pair of verbs in the set (i.e., overlap in unique word-arguments). For clustering, we applied a k-means analysis (Hartigan & Wong, 1979) to the argument-distributions of all verbs in the set, 100 times for 9 different group sizes (from 2-10 groups).

**Method**

**Materials** The set of UP verbs shown in Table 2 were used in the analysis along with their arguments and the frequency with which those arguments occurred with each verb.

**Procedure** Taking all these verbs (both UP and DOWN sets) we created a matrix that recorded all of their arguments and relative frequencies of occurrence. For the *overlap analysis*, we then computed the proportional overlap between the arguments of all pairs of verbs in either set. This was done, first, for all unique-arguments, without regard to frequency, and secondly, using cosine-similarity to take into account arguments' frequencies. We assumed that if more than 60% of verb-B's arguments were covered by verb-A's and if verb-B's coverage of verb-A was asymmetric (i.e., at least 20% lower) then verb-B was a subordinate of verb-A.

For the *clustering analysis*, we computed k-means clustering between the argument-distributions of every verb pair[2]. K-means clustering is an iterative partitional clustering algorithm in which a random set of centroids are chosen to partition the data-points based on their Euclidean distance. Each iteration of the algorithm uses the previous iteration's centroid assignments, the result after *n* runs being a local best-cluster for a given data-point. K-means and other clustering techniques have been applied to a number of NLP and related areas (Cimiano, Hotho & Staab, 2005; Meadche & Staab, 2001). Using k-means clustering, each verb is cast as an 849-word vector with its associated frequency (with a zero-frequency if the word was not used as an argument). Because k-means clustering can produce different results on different runs (based on the random seeding of the centroid locations) we ran it 900 times, 100 runs set at 2-10 groups. As one forces the clustering to generate more groups there is a tendency for the verb-sets to break up into more diverse clusters; though if the similarity is strong between the items, a given group will be maintained and many empty-groups will be produced. Thus, by looking for 10 groups, we are probably biasing the analysis against our predictions, that is, against finding consistent clusters of verbs.

---

[2] K-means clustering was done using the ai4r package available at ai4r.rubyforge.com.

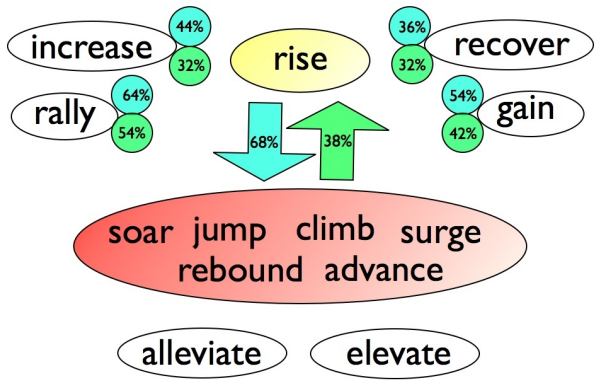

**Figure 2**: Coverage of the main UP verbs; *rise* has asymmetric coverage of the *soar*-to-*advance* subset but has more symmetric even coverage with *increase*, *rally*, and *gain*; *alleviate* and *elevate* seem quite separate in having low coverage of other verbs.

## Results & Discussion

Overall, a clear structure emerges in the argument distributions of the UP set, with *rise* emerging as superordinate-metaphor organising a subset of more specific metaphors, which is substantiated by the clustering analysis.

*Coverage*. Figure 2 summarises the results of the coverage analysis of the UP verbs. It shows that, on average, *rise* covers 68% of the unique-argument distributions of a subset of six verbs (*surge, soar, jump, advance, climb* and *rebound*) and that each verb in this subset is asymmetric in its coverage of *rise* (on average they cover 38% of *rise*). This *soar*-to-*advance* subordinate subset looks like a group of sibling metaphors as they all have quite symmetric coverage of one another (42% versus 48% in either direction[3]).

There are two other identifiable groups in the coverage analysis: a set that could be metaphor-siblings of *rise* and some outliers. The metaphor-siblings to *rise* – *rally, gain, increase*, and *recover* – are variably overlapped by *rise* (36 - 64%) but are more symmetric in their coverage of rise (32 - 54%). They differ from the subordinate-subset in this symmetry; they also have much lower overlaps with the *soar*-to-*advance* subset (on average 25 - 30% in either direction).

The outliers are *elevate* and *alleviate* in that have very little overlap with all of the other verbs in the set, including *rise* (on average <10% in either direction).

*Clustering*. The clustering analysis substantiates some of the structure revealed in the coverage analysis. Over the 900 runs of the k-means analysis, *rise* emerges as a separate group from all the other verbs in 87% of cases. Table 3 shows the 5 most frequent groupings that account for most of the clusters found. The subordinate-subset would fall each time into the rest-category and hence their respective similarity to one another is supported (though, the other verbs are included with them). Though the percentages are low, the only other significant point is that *gain* emerges as a separate group that is related to *jump* and *climb*, a group for which there is a small amount of evidence in the coverage analysis[3].

---

[3] See www.scss.tcd.ie/~gerowa/publications/data/ for full data.

**Table 3**: Top five clusters in k-means analysis of UP-verbs.

| Rank | Cluster Groups | % of Tot. (Freq) |
|---|---|---|
| 1 | rise, rest* | 62% (1451) |
| 2 | rise, gain, rest* | 18% (702) |
| 3 | rise, [climb, gain], rest* | 4% (36) |
| 4 | rise, [jump, climb, gain], rest* | 3% (27) |
| 5 | all-verbs-as-one-group | 2% (18) |

*rest = the remaining verbs in the set

*Summary*. Overall, there is clear evidence from the perspective of coverage and similarity clustering for the pre-eminence of a rise-metaphor as an organizing concept for the UP-metaphors in this finance domain. While the asymmetry in coverage is most marked with the subordinate-subset (*soar, jump, climb, surge, rebound, advance*), there is some evidence for the clustering that *gain* may also organise a *climb-jump* subgroup as well.

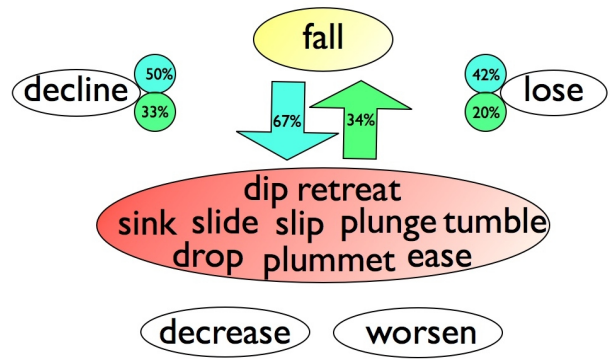

**Figure 3**: Coverage of the main DOWN verbs; *fall* has asymmetric coverage of a 10-verb subset but has more even coverage with *decline* and *lose*; *decrease* and *worsen* which seem quite separate in having low coverage of other verbs.

## Study 2: DOWN Verbs & Metaphors

In this study, we performed the same analysis on the DOWN verbs in the corpus, computing their overlap and the clustering of their respective argument-distributions.

## Method

**Materials** The set of DOWN verbs shown in Table 2.
**Procedure** The procedure applied to the DOWN verbs was identical to that used for the UP verbs.

## Results & Discussion

Looking across the 900 runs of the algorithm (100 clustering runs for groups of 2-10) the DOWN-verb set falls into similar patterns to those found for the UP verbs in both the coverage and the clustering analyses.

*Coverage*. Figure 3 summarises the results of the coverage analysis of the DOWN verbs. It shows that on average *fall*

covers 67% of the unique-argument distributions in a subset of 10 verbs (*sink, tumble, slide, slip, dip, drop, plunge, plummet, retreat*, and *ease*) and that each verb in this subset is asymmetric in its coverage of *fall* (on average they only cover 34% of *fall*'s unique arguments). This *sink-ease* subordinate-subset looks like a group of sibling metaphors as they all have quite symmetric coverage (on average 41% versus 42% in each direction[4]).

There are two other identifiable groups in the coverage analysis: a pair that may be siblings of *fall* and some outliers. The metaphor-siblings of *fall* – *decline* and *lose* – have larger overlaps with *fall* (42% and 50%, respectively) but are not as marked in their asymmetry as the 10-verb subset (33% and 20%, respectively). They differ from the 10-verb subset in having less overlap and lower asymmetry; they also have lower overlaps with the 10-verb subset (25% v 30% in either direction[4].

The outliers are *worsen* and *decrease* that have very little overlap with all of the other verbs in the set, including *fall* (on average <20% for most).

*Clustering*. The clustering analysis substantiates some of the structure revealed in the coverage analysis. Over the 900 runs of the k-means analysis, *fall* emerges as a separate group from all the other verbs in 89% of cases. Table 4 shows the 5 most frequent clusters that account for the majority of those found. The 10-verb subset is part of the rest-category and hence their respective similarity to one another is supported (though, other verbs are included with them). Though the percentages are low, the only other significant point is that *lose* emerges as a separate group, that is related to *drop* and *slip* (there is some, but not a lot of evidence for this in the coverage analysis[4]).

**Table 4**: Top five k-means clusters of DOWN-verbs.

| Rank | Cluster | % of Total (Freq) |
|---|---|---|
| 1 | fall, rest* | 71% (639) |
| 2 | fall, [drop, lose], rest* | 11% (99) |
| 3 | fall, lose, rest* | 5% (45) |
| 4 | fall, [slip, drop, lose], rest* | 2% (18) |
| 5 | all-verbs-as-one-group | 2% (18) |

*rest = the remaining verbs in the set

*Summary*. Overall, there is clear evidence from the perspective of the coverage analysis and similarity clustering for the pre-eminence of *fall* as an organising concept for the DOWN metaphors in this domain. While the asymmetry in coverage is most marked with the 10-verb subset (*sink, tumble, slide, slip, dip, drop, plunge, plummet, retreat*, and *ease*) there is some evidence from the clustering that *lose* may also organise a *slip-drop* subgroup.

# General Discussion

What is clear from these studies is that something can be said about the conceptual organization of metaphors from a corpus analysis of their argument distributions. Our main conclusion is that these metaphors are organised hierarchically, with a definite organizing superordinate metaphor with subordinate groupings of sibling metaphors. The existence of such an ontology of metaphors is not surprising from a Lakoffian viewpoint. The main novelty of the current paper is that it provides an empirical / computational method for identifying such metaphoric structures.

In both studies, a distinctive over-arching metaphor is found; *rise* and *fall*, both of which are simple forms of the MORE IS UP / LESS IS DOWN metaphor. In each case, deeper inspection finds clusters of less pervasive metaphors, such as *climbing* and *falling* as well as *gaining* and *dropping*. The clustering of such metaphors in UP as well as DOWN verbs is not surprising, especially in the domain of finance, in which the movement of markets, indices, prices, and other quantitative metrics are often made sense of using such conceptual metaphors. Assuming the Lakoffian view, that conceptual metaphors are different than the linguistic forms they take, we are forced to admit their conceptual structure is partially obfuscated. The current paper, however, offers a method of discerning the structure of metaphors by way of a statistical analysis of language.

In a separate paper (Gerow & Keane, 2011b) we present a similar analysis of the results of a psychological experiment where people were asked to match antonyms of metaphoric verbs. A comparison of UP- and DOWN-verbs, explicitly paired, presents an approach to finding metaphor-relationships in a strict semantic relationship: antonymy. Similar to this paper, a distributional analysis is shown to uncover the deeper structure of metaphors to support the Lakoffian theory that they exhibit conceptual form. This form is made explicit through a corpus analysis of the language used to instantiate the metaphors. A computational approach to metaphor is analogous to a distributional semantics approach to meaning, such as LSA and other vector space models (Landauer & Dumais, 1991; Turney & Pantel, 2010). The analyses in both papers bridge a gap between traditional semantics and a modern understanding of metaphor, specifically, Lakoff's theory of conceptual metaphor.

We have shown how a corpus-based analysis of verbs substantiates the proposal that knowledge-structuring metaphors exhibit hierarchical organization. This work is part of a growing field of vector space semantic models (see Turney & Pantel, 2010 for a review) and extends similar approaches to address the structure of metaphor. Crucially, it combines a modern view of metaphor, which has remained largely theoretical, with statistically-derived models of meaning. Metaphors are slippery creatures in linguistics, and are central to the way we think — bridging these fields is critical to cognitive linguistics as well as lexical semantics.

## Acknowledgements

This work was carried out as part of a self-funded MSc in the Cognitive Science programme at the University College Dublin by the first author. Thanks to Trinity College Dublin and Prof. K. Ahmad for support and supervision of the first author's PhD during the preparation of this paper. Thanks to K. Hadfield and four anonymous reviewers for valuable suggestions.